%% file: main.tex
\documentclass{article}

\usepackage[preprint]{neurips_2020} 

\usepackage[utf8]{inputenc} 
\usepackage[T1]{fontenc}    
\usepackage[hidelinks]{hyperref}       
\usepackage{url}            
\usepackage{booktabs}       
\usepackage{amsfonts}       
\usepackage{nicefrac}       
\usepackage{microtype}      
\usepackage{wrapfig}
\usepackage{graphicx}
\usepackage{subfig} 

\input{macros}

\title{Domain Adaptation with Morphologic Segmentation}

\author{%
\begin{tabular}{lll}
{\bf Jonathan Klein} & \textnormal{University of Bonn} & \texttt{kleinj@cs.uni-bonn.de} \\
{\bf S\"oren Pirk} & \textnormal{Google Brain} & \texttt{pirk@google.com} \\
{\bf Dominik L.~Michels} & \textnormal{KAUST} & \texttt{dominik.michels@kaust.edu.sa}\\
\end{tabular}
}

\begin{document}

\maketitle

%

\input{00-abstract}

\input{01-introduction}

\input{02-related_work}
\input{03-method}

\input{04-implementation}
\input{05-results}
\input{06-conclusion}

\section*{Acknowledgements}\vspace{-4mm}
This work has been supported by KAUST under individual baseline funding.

\clearpage
\input{07-impact}
\bibliographystyle{unsrt}
\bibliography{main}

\end{document}

%% file: macros.tex
\usepackage{xcolor}

\newcommand{\smallsection}[1]{\vspace{-2mm}\section{#1}\vspace{-4mm}}



%% file: 00-abstract.tex
\begin{abstract}

We present a novel domain adaptation framework that uses morphologic segmentation to translate images from arbitrary input domains (real and synthetic) into a uniform output domain. Our framework is based on an established image-to-image translation pipeline that allows us to first transform the input image into a generalized representation that encodes morphology and semantics -- the edge-plus-segmentation map (EPS) -- which is then transformed into an output domain. Images transformed into the output domain are photo-realistic and free of artifacts that are commonly present across different real (e.g.~lens flare, motion blur, etc.) and synthetic (e.g.~unrealistic textures, simplified geometry, etc.) data sets. Our goal is to establish a preprocessing step that unifies data from multiple sources into a common representation that facilitates training downstream tasks in computer vision. This way, neural networks for existing tasks can be trained on a larger variety of training data, while they are also less affected by overfitting to specific data sets. We showcase the effectiveness of our approach by qualitatively and quantitatively evaluating our method on four data sets of simulated and real data of urban scenes.

\end{abstract}

%% file: 01-introduction.tex
\smallsection{Introduction}

\label{sec:introduction}
Deep neural networks have shown unparalleled success on a variety of tasks in computer vision and computer graphics, among many others. To enable this paramount performance a majority of approaches relies on large and diverse data sets that are difficult to establish. Especially, for image-to-image tasks, where an image is transformed into a different representation (e.g. semantic segmentation), labels are extremely difficult to obtain. Furthermore, even large and well established data sets such as KITTI~\cite{Geiger2013IJRR} and Cityscapes~\cite{Cordts2016} are biased and do not contain enough variance to allow for training models that are robust and generalize well. Synthetic data generated with modern rendering and simulation algorithms may provide a solution to this problem. However, the disparity between real and synthetic data often limits using generated data to train models so as to operate on real data. A number of approaches aim to explicitly overcome this limitation by modeling the sim-to-real transfer based on Generative Adversarial Networks (GANs) that transform synthetically generated images into images with more photo-realistic visual traits, such as shadows, highlights, or higher frequency textures~\cite{Wang2018,46717,8954361,8237506,Yi2017DualGANUD}. However, while some of the existing approaches show impressive results, it remains challenging to faithfully reconstruct the full spectrum of details in real images.


In this paper, we introduce a novel framework that translates images from various data sources to a unified representation. Synthetically generated images are enhanced, while the details of real photographs are reduced. Our goal is to establish a preprocessing step that translates data from different sources and of different modalities (real, synthetic) into a unified representation that can then be used to train networks for computer vision downstream tasks, such as classification, object detection, or semantic segmentation. This has the advantage that the downstream network can be trained on less complex data and -- in turn -- usually with a less complex architecture.  Our  image-to-image translation approach combines a state-of-the-art semantic segmentation algorithm~\cite{Soria2019} with an edge detection algorithm~\cite{Chen2018} to generate edge-plus-segmentation (EPS) maps from arbitrary input images. We refer to this intermediate representation as \textit{morphologic segmentation}. Instead of directly working on RGB images our GAN only uses the EPS maps to generate output images. Hence, our method is agnostic to the style and details of the input images. Our generator network builds upon the state-of-the-art \textit{pix2pixHD} algorithm \cite{Wang2018}. Instead of using contour maps, we employ edge maps that also show internal object edges to provide further guidance for the image generation process. Moreover, we use a progressive learning scheme to achieve high-resolution output images and reduced training time.

To show that the automatic generation of training data is feasible and that EPS maps serve as a meaningful intermediate representation, for generating photo-realistic images with plausible amounts of detail, we train a generator on image data collected from YouTube videos. Once trained, the generator can then be used for the translation of images from various sources, including CARLA~\cite{Dosovitskiy17,hendrycks2019anomalyseg}, Cityscapes~\cite{Cordts2016}, FCAV~\cite{Johnson2017}, and KITTI~\cite{Geiger2013IJRR} as well as across different modalities. To validate the impact of EPS maps, we run ablation studies and measure common metrics for semantic segmentation. 

In summary, our contributions are (1) we introduce a novel framework for translating images from different sources into a unified representation; (2) we introduce \emph{edge-plus-segmentation} (EPS) maps as a novel representation for training image-to-image translation networks so as to operate on inputs of different sources and modalities; (3) we evaluate our approach on a range of different data sources and perform ablation studies using established metrics.




\begin{figure}[t]
\center
\includegraphics[width=1\linewidth]{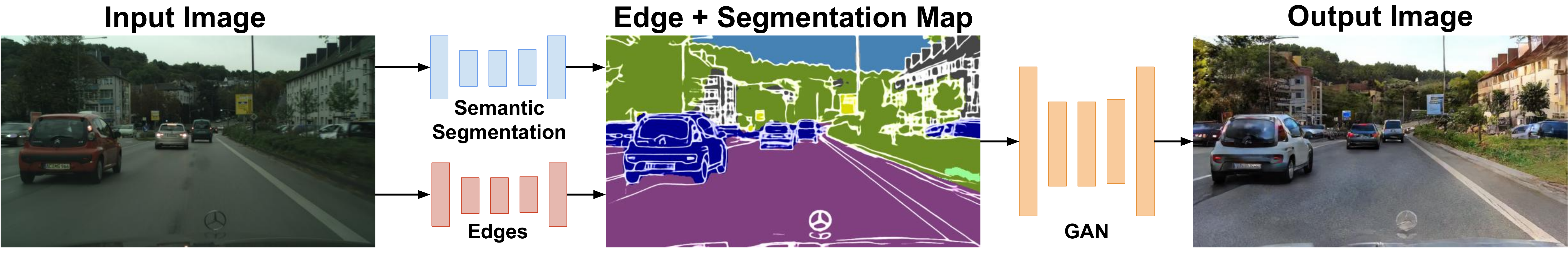}
\vspace{-6mm}
\captionof{figure}{
An input image is transformed into the output image using our framework: we first generate an EPS map as intermediate representation to then generate the output image in the target domain with an image-to-image translation GAN. The input image shown here for illustration is taken from the Cityscapes data set and object classes of the EPS map are visualized based on the color scheme used therein.}
\vspace{-6mm}
\label{fig:teaser}
\end{figure}

%% file: 02-related_work.tex
\smallsection{Related Work}
\label{sec:related_work}

\textbf{Generative Adversarial Networks.} GANs~\cite{Goodfellow2014} provide a powerful means for image generation by training a generator network so that it generates images that are indistinguishable from a target distribution. Furthermore, it has been recognized that GANs can be a applied to a wide range of image synthesis and analysis problems, including image-to-image translation~\cite{Isola2017,Mirza2014,Ronneberger2015,Long2015} or -- more generally -- image generation~\cite{pmlr-v70-arjovsky17a,Reed2016a,Wang2016,76256ca3b51d4ed79ba46d0a3c30e37d}, image manipulation~\cite{Zhu2016GenerativeVM}, object detection~\cite{8099694}, and video generation~\cite{Mathieu2015DeepMV,10.5555/3157096.3157165}. As training on large image resolutions easily becomes infeasible, many approaches aim to improve the output resolution, often based on stacked architectures~\cite{Wang2018,8099685,Zhang2017,Karras2018,Ledig2017}. Similar to many of the existing approaches, we rely on a GAN-based pipeline to convert images to a common target domain. 

\textbf{Image-to-Image Translation.} A number of methods use GANs for image-to-image translation. To this end, Isola et al.~\cite{Isola2017} present \textit{pix2pix}, the first approach that takes inspiration from established networks~\cite{Ronneberger2015,Long2015} to transfer images between domains. Although pix2pix can be applied to a wide variety of domains, it is limited by the supported output resolution and it requires a large amount of training data. Various approaches address this and related problems. Zhu et al.~\cite{Zhu2017a} introduce cycle-consistency loss: images are first translated to the target domain and then back to the source domain; loss is measured over the generated and the input image. Cycle consistency allows generating impressive results across a large variety of image domains~\cite{Zhu2017b}. Liu et al.~\cite{Liu2019} present a method that can generate images from previously unseen classes from only a few examples. Similarly, \textit{Disco GANs} learn cross-domain relations and can transform instances of one class to similar looking instances of another class without requiring extra labels \cite{Kim2017}. Huang et al.~\cite{Huang2018} transfer an input image into a shared content space and a class specific style space, which allows recreating images of different classes in an unsupervised manner. Chen et al.~\cite{Chen2017} propose cascaded refinement networks, which drastically increases the output resolution and feature quality. On a slightly different trajectory, Gatys et al.~\cite{Gatys2015} introduce a neural style-transfer network, which is able to transfer among various artistic styles of images; for a comprehensive overview we refer to Jing et al.~\cite{Jing2019}. While style transfer networks primarily focuses on the visual appearance of images, they do not capture any semantic properties, which limits using these algorithms for many types of image-to-image transfer. 


\textbf{Domain Adaptation.} Methods for domain adaptation use neural networks to more generally model the domain shift of source and target domains by either converting one domain into the other or by lifting source and target domain into a shared domain. To this end, Tzeng et al.~\cite{7410820} use a CNN-based architecture to model domain invariance for domain transfer, while Long et al.~\cite{10.5555/3045118.3045130} propose Deep Adaptation Network (DAN) to match domain distributions in a Hilbert space. Murez et al.~\cite{8578571} propose a method for the domain adaptation by more explicitly constraining the extracted features of the encoder network, which enables them to model the domain shift of unpaired images. Moreover, it has been recognized that adversarial losses can be leveraged for modeling the domain shift. This includes GANs that model for unequally labeled data of source and target domains~\cite{10.5555/3294771.3294787}, for improving the realism of simulated images~\cite{8099724}, for unsupervised training setups~\cite{Sankaranarayanan2017GenerateTA}, and more general approaches that combine discriminative modeling and weight sharing with an adversarial loss~\cite{8099799}. Similar to the existing approaches our goal is to model the domain shift of different source domains so as to lift samples into a common target domain. However, unlike them we use EPS maps to translate images of a variety of source domains into a common target domain.  

%% file: 03-method.tex
\smallsection{Method}

Our method consists of two main steps: first, we introduce morphologic segmentation as the combination of semantic segmentation and edge detection. Specifically, we generate an edge-plus-segmentation (EPS) map as a representation for input images. In a second step, we generate an output image by only synthesizing it from an EPS map. In the following we describe how we generate EPS maps and how we use them to synthesize photo-realistic images with a GAN.


\begin{figure}
	\centering
	\includegraphics[width=\columnwidth]{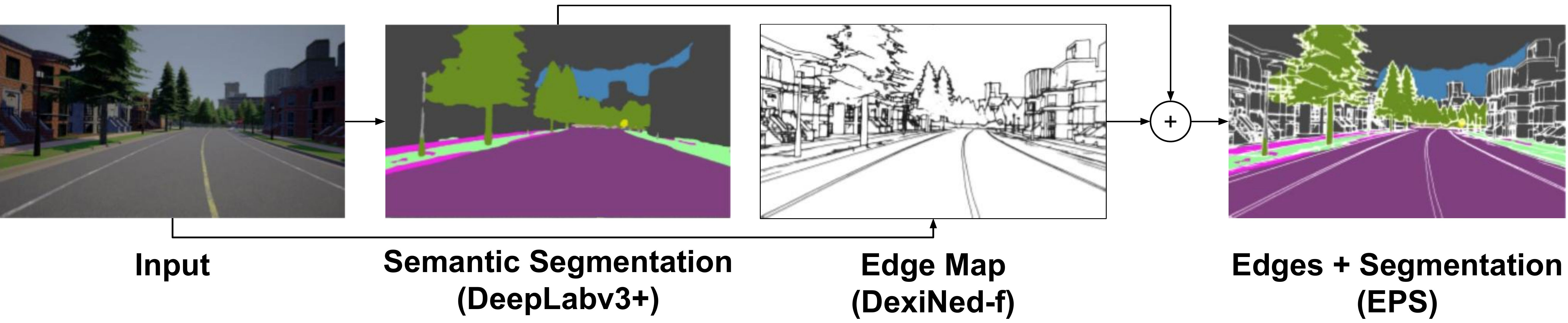}
	\vspace{-5mm}
	\caption{\label{fig:eps_creation} EPS maps are generated by combining a semantic segmentation map and an edge map generated by two state-of-the-art networks.}
	\vspace{-7mm}
\end{figure}

\textbf{EPS Map Generation.} We introduce EPS maps as an abstract intermediate representation of an image that encodes semantic information of objects as well as their morphology defined as edges. We argue that a common semantic segmentation map is insufficient as it does not provide intra-object details. Therefore, we extend semantic segmentation maps with edges, which encode more intricate geometric details. We aim to devise a representation for images that  abstracts away defining details such as textures, lighting conditions, or the style of an image, that may be unique to a specific data set. Our goal is to train a generator network so that it can synthesize an image with uniform, yet complex visual traits, by only using the EPS map. 


Edge detection as well as semantic segmentation have been active areas of research for many years. We use state-of-the-art approaches for semantic segmentation (DeepLabv3)~\cite{DBLP:journals/corr/ChenPSA17} and edge detection (DexiNed-f)~\cite{soria2019dense}. Please note that both networks are in principal interchangeable without affecting the rest of the pipeline. DeepLabv3 is the state-of-the-art extension of DeepLabv1~\cite{DBLP:journals/corr/ChenPKMY14} and DeepLabv2~\cite{7913730}. In the previous versions, the input image is processed by the network using atrous convolution increasing the receptive field while maintaining the spatial dimension of the feature maps followed by bilinear interpolation and fully connected conditional random field (CRF) postprocessing taking context into account. DeepLabv2 employs ResNet~\cite{7780459} and VGGNet~\cite{DBLP:journals/corr/SimonyanZ14a} while DeepLabv1 only uses VGGNet. DeepLabv2 further employs atrous spatial pyramid pooling (ASPP) enabling the network to encode multi-scale contextual information. In DeepLabv3, ASPP is further augmented with features on the image-level encoding global context information and further boosting performance. It outperforms its predecessors even without the CRF postprocessing. Chen et al.~\cite{Chen_2018_ECCV} then further extended DeepLabv3 to DeepLabv3+ mainly by adding a decoder module refining the segmentation results especially along object boundaries.

DexiNed-f is an edge detector devised by Soria et al.~\cite{soria2019dense} providing fused edge maps. It is based on holistically-nested edge detection (HED)~\cite{DBLP:journals/corr/XieT15} enabling holistic image training and prediction while allowing for multi-scale and multi-level feature learning. Its general architecture is inspired from the Xception network~\cite{DBLP:journals/corr/Chollet16a}.

Figure~\ref{fig:eps_creation} shows how we compute EPS maps. The input image is passed to DexiNed-f to produce a grayscale image of the edges, while we use DeepLabv3+ to generate the segmentation map. Both maps are combined by pixel-wise addition to create the EPS map. 
We argue that EPS maps reduce the number of input dimensions, while providing enough meaningful details to reconstruct photo-realistic images.

\textbf{Image Synthesis.} Our image generator builds upon \textit{pix2pixHD}~\cite{Wang2018} as an extension of pix2pix~\cite{Isola2017} overcoming the lack of fine details and unrealistic textures. In this regard, the authors introduced a novel adversarial learning objective increasing robustness together with optimized multi-scale architectures of the generator and the discriminator networks. As in pix2pix, loss functions are specifically tailored to the input domain.


The original pix2pixHD implementation supports two different encodings for the segmentation maps: a one-hot encoding for each class, or color-coded RGB maps with different colors representing different classes. Due to the encoding of the EPS maps, the latter one is used. In the original implementation an object instance map can be provided along the segmentation map. Since the edge information from the EPS maps already encode similar information (however, edge maps have not necessarily closed object boundaries and also contain inner edges), we do not make use of this additional input channel. While pix2pixHD is used without modifying its architecture, the type of input information and its encoding is notable different from the original proposed.


%% file: 04-implementation.tex
\smallsection{Implementation and Training}

For generating EPS maps we use pretrained instances of DeepLabv3+ and DexiNed-f. 
Depending on the supported classes and used training data, different versions of DeepLabv3+ are available. We use the Xception71 variant, which is trained on the Cityscapes~\cite{Cordts_2016_CVPR} data set, which outputs segmentation classes commonly found in the Cityscapes images. While DexiNed-f is very versatile and performs well on a large set of input domains, semantic segmentation is more problem tailored. Thus, when executing our framework on different image types, the trained DeepLabv3+ network would need to be exchanged. Since DeepLabv3+, DexiNed-f, and pix2pixHD are implemented in Python, using either TensorFlow or PyTorch, our EPS pipeline is completely written in Python.

\textbf{Training Data.} The manual labeling of data is a time consuming and often tedious process ~\cite{Johnson2017}. Therefore, to emphasize the independence from manually labeled training data, we abstain from using a classic data set. Instead, we use a large set of YouTube videos (according to fair use guidelines) showing urban scenes from the driver's perspective. This allows us to cover a wider variety of scenarios compared to what is contained in any of the currently available data sets, that are shown in Figure~\ref{fig:images_examples}. In total, 98 different videos corresponding to about 45~hours of video material were used. The videos have been selected using appropriate hashtags ensuring that most scenes are showing European roads at daytime summer days, providing good weather conditions. From these videos, around 101\,000 frames were randomly extracted and used for training. Many videos contain unsuitable scenes. Therefore, we manually removed video segments prior to randomly selecting frames for training.

Our data set is further augmented using a combination of three simple steps: first, each image is rotated around a randomly selected center by $\pm 7^\circ$. A random cropping of up to half of the image is then applied in each dimension. Finally, we scale the image by $\pm20\%$, while preserving the aspect ratio. After applying these augmentation steps, the number of training artifacts is greatly reduced.



\textbf{Progressive Training.} We train generator and discriminator by progressively increasing the image resolution in four steps (each doubling the resolution); starting with a resolution of $128\times72$ up to a final resolution of $1024\times576$. Increasing the resolution results in finer details as the training progresses. Accordingly, the learning rate of $\eta=0.0001$ used in the first step is decreased with increasing resolution to $\eta=0.00001$ in the second and third steps. However, in the final step, the initial learning rate $\eta=0.0001$ is applied again in order to avoid overfitting, which results in improved stability~\cite{DBLP:journals/corr/abs-1710-10196}. 

%% file: 05-results.tex
\smallsection{Evaluation}

\begin{figure}
\centering
\includegraphics[width=0.99\columnwidth]{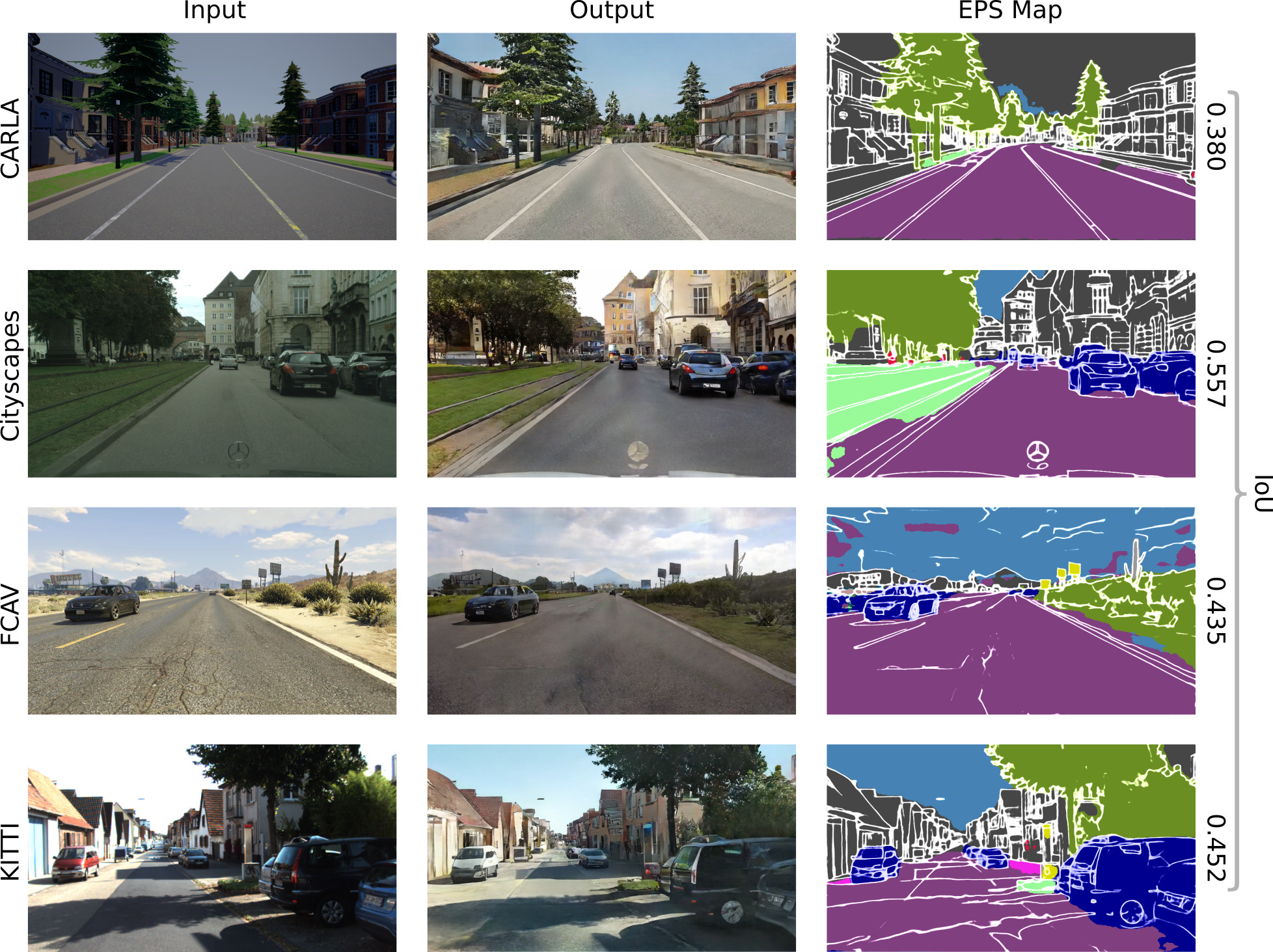}
\caption{\label{fig:images_examples}Examples of synthesized images from various input sources CARLA, Cityscapes, FCAV, and KITTI comprising real and rendered data. The EPS maps are extracted from the input images and then used as the only input to synthesize the output images.}
\vspace{2mm}
\centering
\includegraphics[width=0.49\columnwidth]{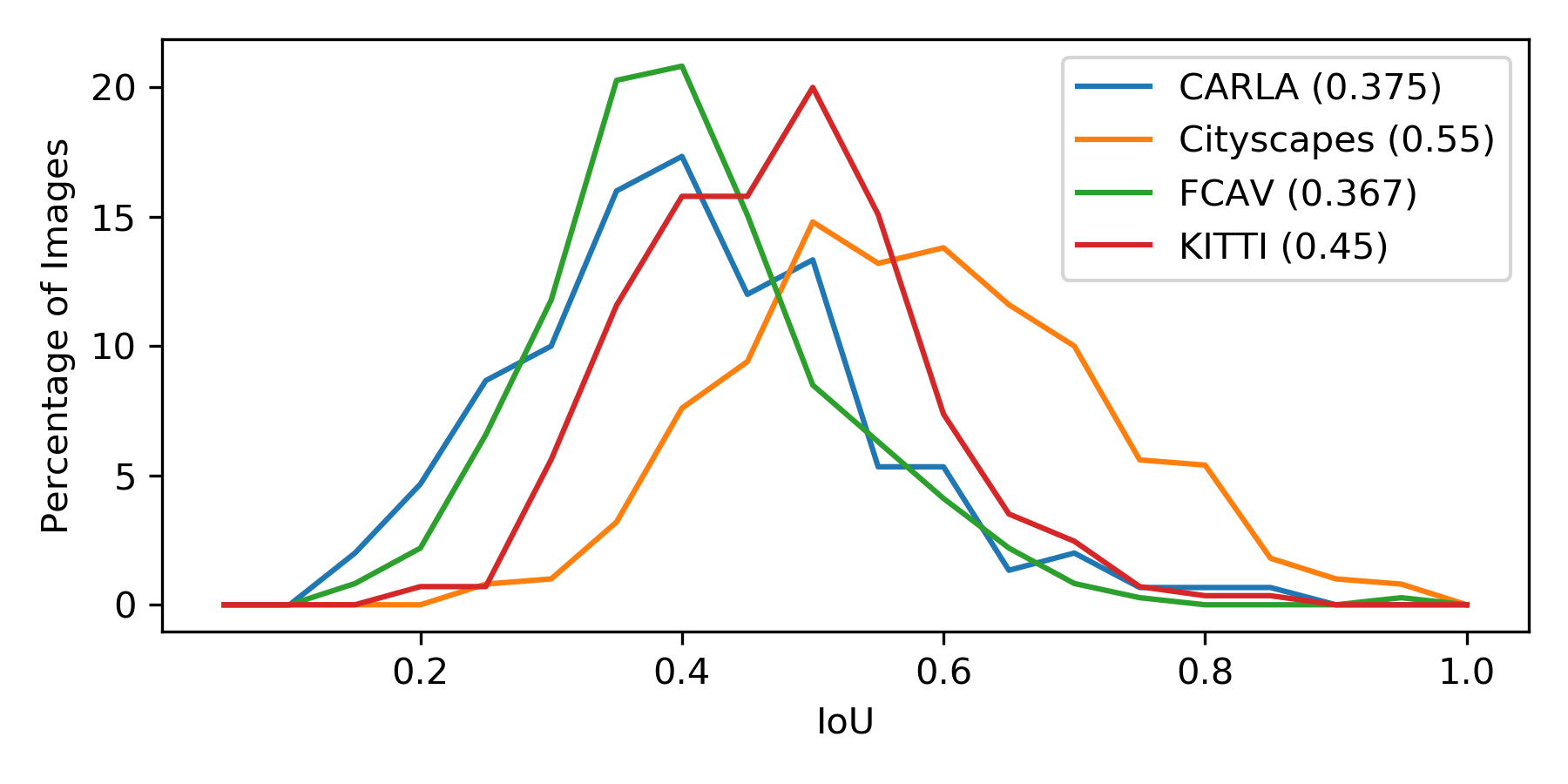}
\hfill
\includegraphics[width=0.49\columnwidth]{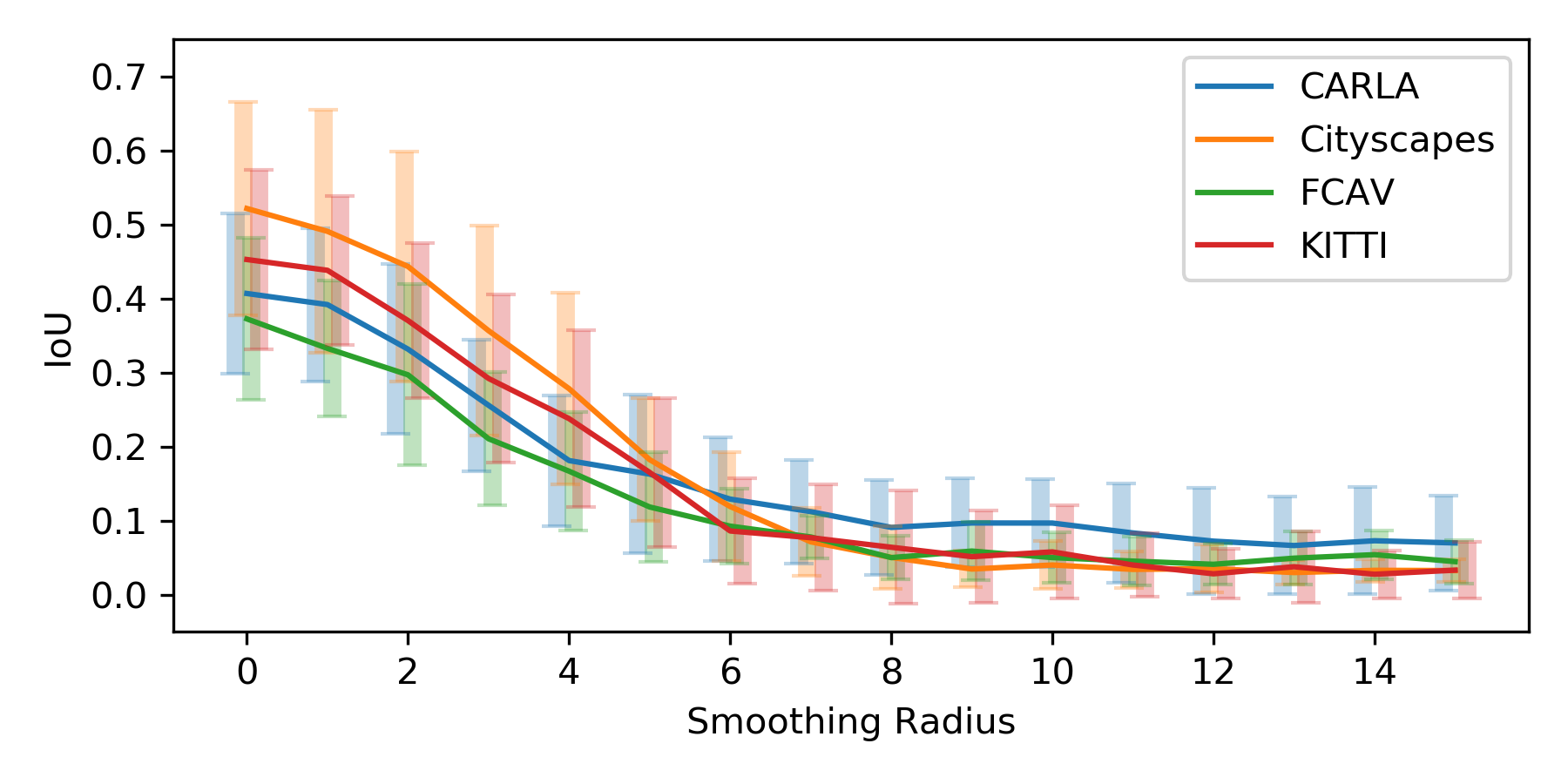}
\vspace{-3mm}
\caption{\label{fig:class_iou}Illustration of the $\mathsf{IoU}$ scores per class considering CARLA, Cityscapes, FCAV, and KITTI (left), and the dependence of the $\mathsf{IoU}$ score on different smoothing radii (right).}
\vspace{-2mm}
\end{figure}

Examples of synthesized images from various input sources are shown in Figure~\ref{fig:images_examples}. While the input images across the different data sets show a number of heterogeneous visual features, such as varying lighting condition or diverse environments, the translated images only show consistent visual traits. 
For example, while many of the input images taken from the FCAV data set show a desert environment, its corresponding output images show features of more common environments. Furthermore, it can be observed that synthetic input images are enhanced with more photo-realistic details. In contrast, images of real data sets are showing less artifacts commonly present in photos (e.g.~exceedingly strong contrasts, hard shadows, gray cast, etc.). The disparity between real and synthetic data is significantly reduced.

We follow an established evaluation protocol for image-to-image translation~\cite{Wang2018,Isola2017,Zhu2017a}. The quality of our results is evaluated by computing semantic segmentation maps of the output images and by comparing how well the resulting segmentation maps match the corresponding segmentation of the input images. This can easily be quantified by computing the mean intersection-over-union, $\mathsf{IoU}\in[0,1]\subset\mathbb{R}$, score. Figure~\ref{fig:images_examples} lists the $\mathsf{IoU}$ score for the different cases presented therein. The scores range from $\mathsf{IoU}=0.380$ to $\mathsf{IoU}=0.557$ and can be considered as average values. A representative cross section of our results is shown here. Moreover, the distributions of the $\mathsf{IoU}$ scores per class are shown in Figure~\ref{fig:class_iou} (left). For this evaluation, 150 images from CARLA, 500 images from Cityscapes, 365 images from FCAV, and 285 images from KITTI data set were analysed. DeepLabv3+ for the semantic segmentation was employed as in our whole image-to-image translation pipeline. This likely causes the effect that significantly higher scores are usually obtained for the Cityscapes inputs since DeepLabv3+ is trained on the Cityscapes data set~\cite{Chen_2018_ECCV}. 

\begin{figure}
\centering
\includegraphics[width=0.9\columnwidth]{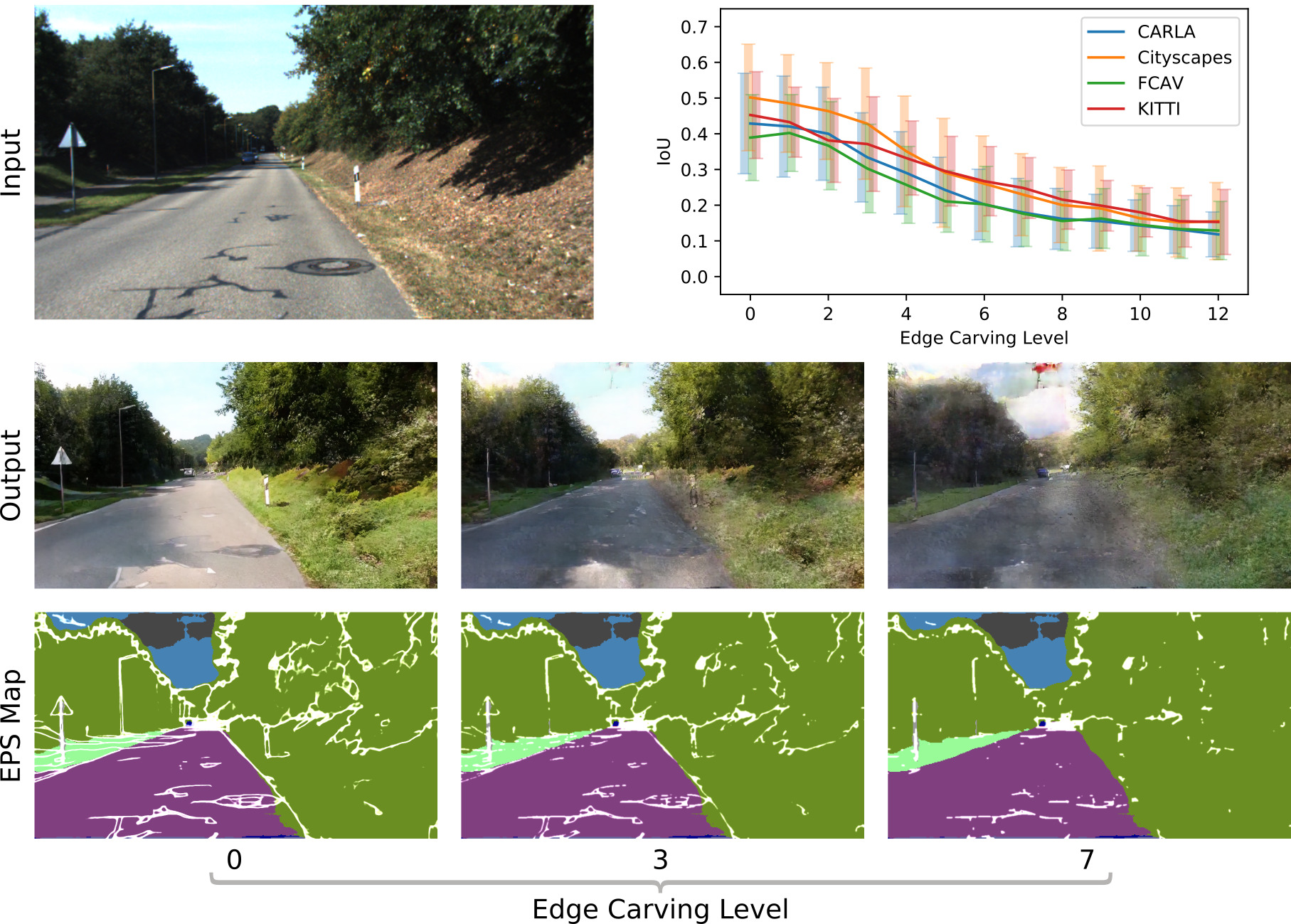}
\caption{\label{fig:edge}Illustration of the effect caused by different edge carving levels and the corresponding $\mathsf{IoU}$ score. The input image shown here for illustration is taken from the KITTI data set.}
\vspace{-2mm}
\end{figure}

\begin{figure}
\centering
\includegraphics[width=0.9\columnwidth]{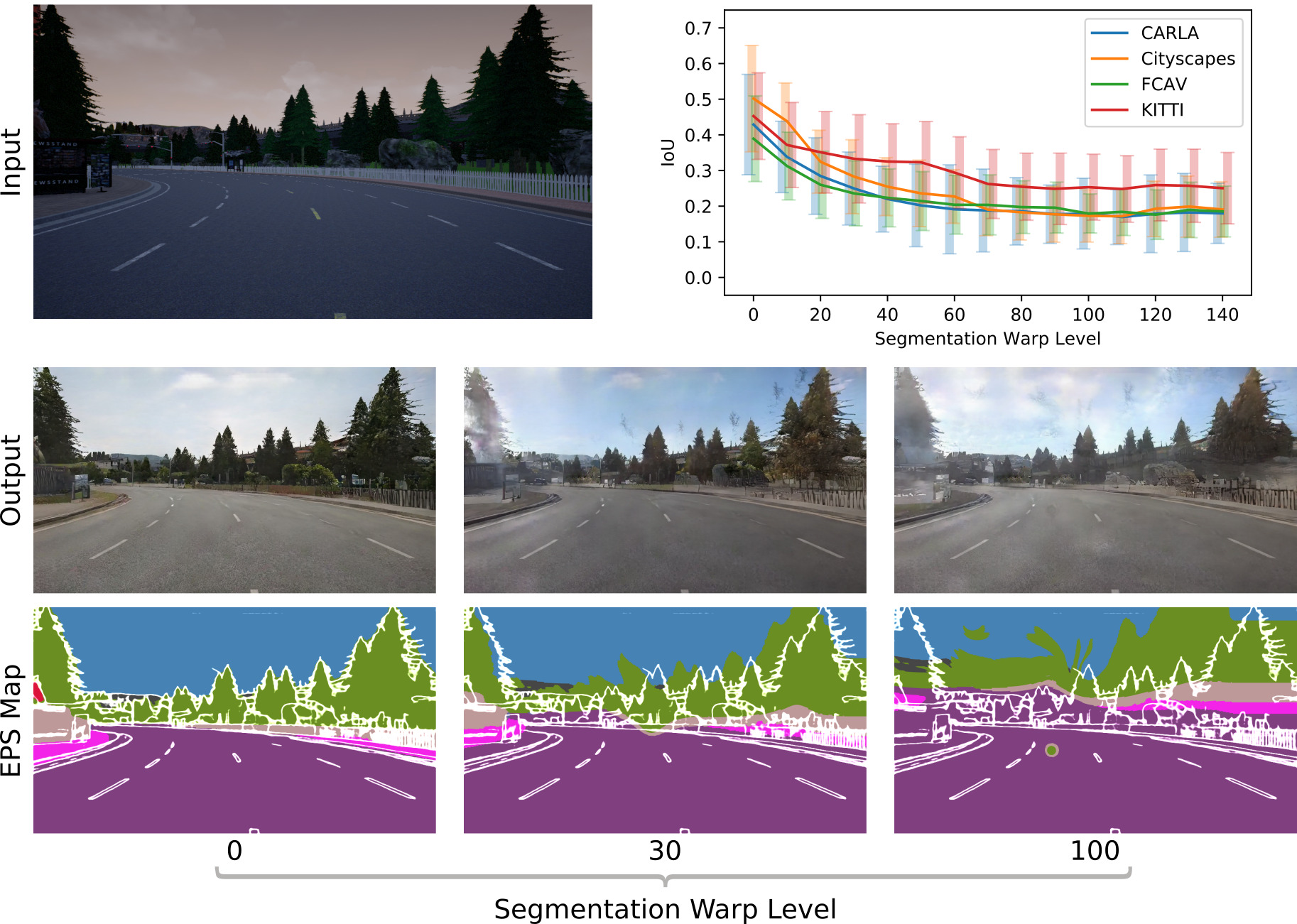}
\caption{\label{fig:segmentation}Illustration of the effect caused by warping semantic segmentations using different intensities, and the corresponding $\mathsf{IoU}$ score. The input image shown here for illustration is taken from the CARLA data set.}
\end{figure}

\begin{figure}
\centering
\includegraphics[width=0.93\columnwidth]{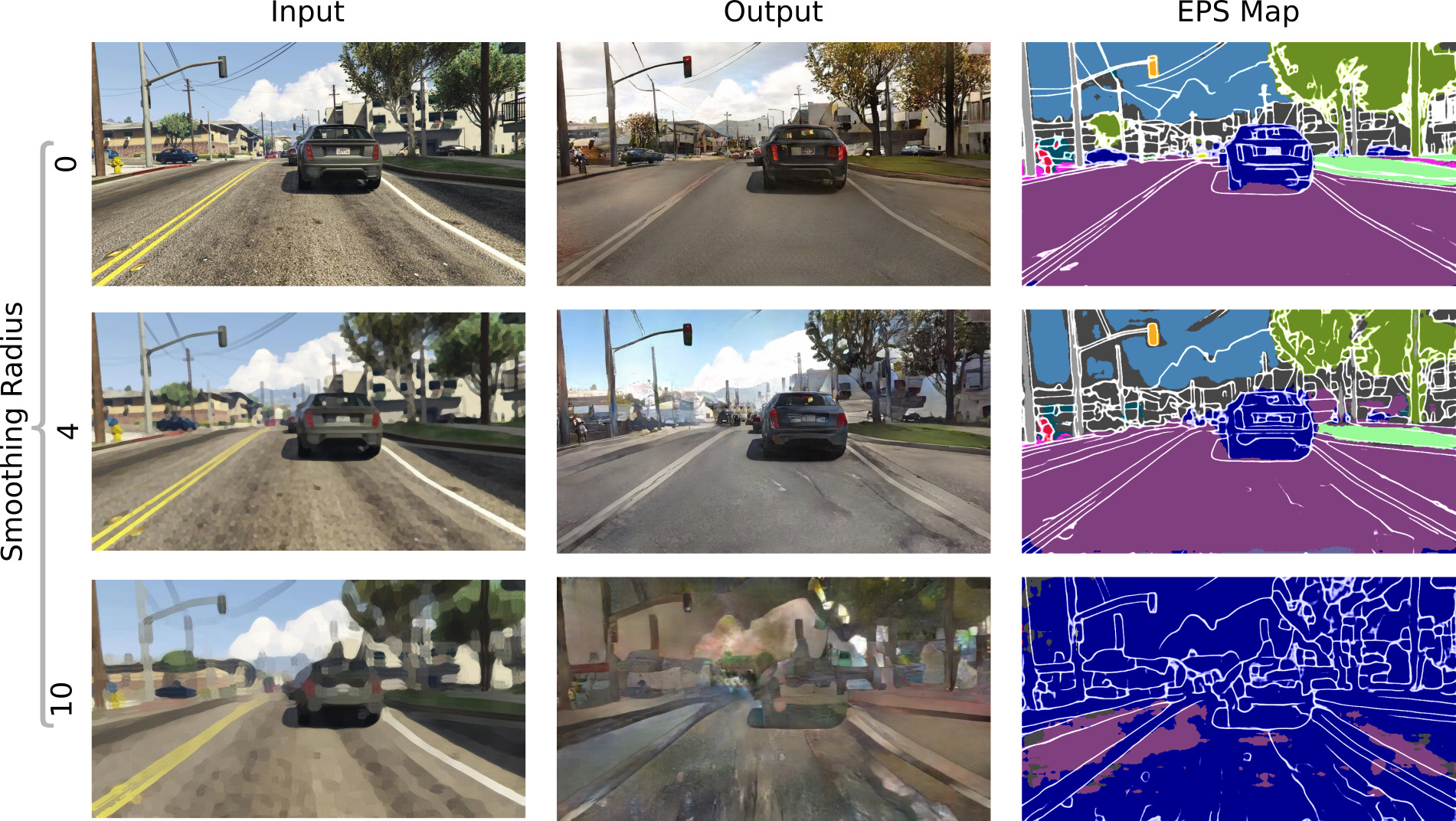}
\vspace{-1mm}
\caption{\label{fig:smoothing}Illustration of the effect caused by smoothing the input image. A Kuwahara filter is applied with different radii. The input image is taken from the FCAV data set.}
\vspace{-6mm}
\end{figure}

We carry out three ablation studies providing further evaluations of our approach. In particular, we systematically analyze the impact of quality of the input data, the availability of accurate edge details, and the quality of the segmentation. In this regard, we analyzed a reduced data set comprising $16$ images of each input source CARLA, Cityscapes, FCAV, and KITTI since each image produced a series of new images as explained in the following. Figure~\ref{fig:smoothing} illustrates the effect caused by smoothing the input image. Specifically, a Kuwahara filter~\cite{Kuwahara1976} is applied using different radii $r$  to control the smoothing intensity. We quantify the effect of smoothing by measuring $\mathsf{IoU}$ scores between segmentations of input and output images for different smoothing intensities. This is illustrated for the different classes in Figure~\ref{fig:class_iou} (right). It can be observed that in the case of real data (Cityscapes and KITTI), the $\mathsf{IoU}$ scores are initially above the ones of synthetic data (CARLA and FCAV), while the scores are getting more similar with increasing smoothing radius. As the general quality of images decreases, the domain disparity between real and synthetic data is reduced. This study also shows significant robustness of our approach with respect to image quality as illustrated in Figure~\ref{fig:smoothing} (middle row) demonstrating the effect of a relatively strong Kuwahara smoothing with $r=4$. The smoothed input image is completely blurred, while the corresponding output image contains fine traits such as structural and texture details of the vegetation located between the street and the sidewalk on the right. Thus, our approach can potentially enable the use of additional data sources for training downstream tasks.

Moreover, we investigate the importance of the accuracy of edges and segmentation with respect to the output quality. Edges are systematically carved as illustrated in Figure~\ref{fig:edge}. Since we observed that most edges already show relatively small gradients, we threshold pixels at $64\%$ intensity in each step followed by a smoothing procedure with a $1$px kernel. This can be repeated to allow for different edge carving levels. The $\mathsf{IoU}$ scores between segmentations of input and output images for different levels are measured further underlying our previous observations. Our image-to-image transformation is robust towards a decrease of image quality while domain disparity between real and synthetic data is reduced. We carry out a similar ablation study by systematically reducing the quality of the semantic segmentation instead of edge quality. This is illustrated in Figure~\ref{fig:segmentation} and consistent with the previous observations. In this regard, segmentation maps are systematically warped to reduce their quality by manipulating control points. Fine details are first warped by shifting control points by distances sampled from a normal distribution with a standard deviation corresponding to the warp level. The result is then warped by a single point along a distance proportional to the warp level concluding a distorting process affecting multiple scales.\footnote{The resolution of our segmentation maps is given by $1024\times576$. A $7\times5$ grid is used in which the inner points are shifted. The result is further transformed by shifting the center point in a random direction by a distance of twice the warp level.} The analysis shows that our approach is more robust with respect to a loss of segmentation quality compared to edge quality. In other words, even if the segmentation quality is rather low, given appropriate edges usually available, proper results can still be achieved.

Since semantic segmentation is a less stable task than edge detection (for instance in Figure~\ref{fig:smoothing} the
\begin{wrapfigure}{r}{0.65\textwidth}
  \vspace{-8pt}
  \begin{center}
 \includegraphics[width=0.65\textwidth]{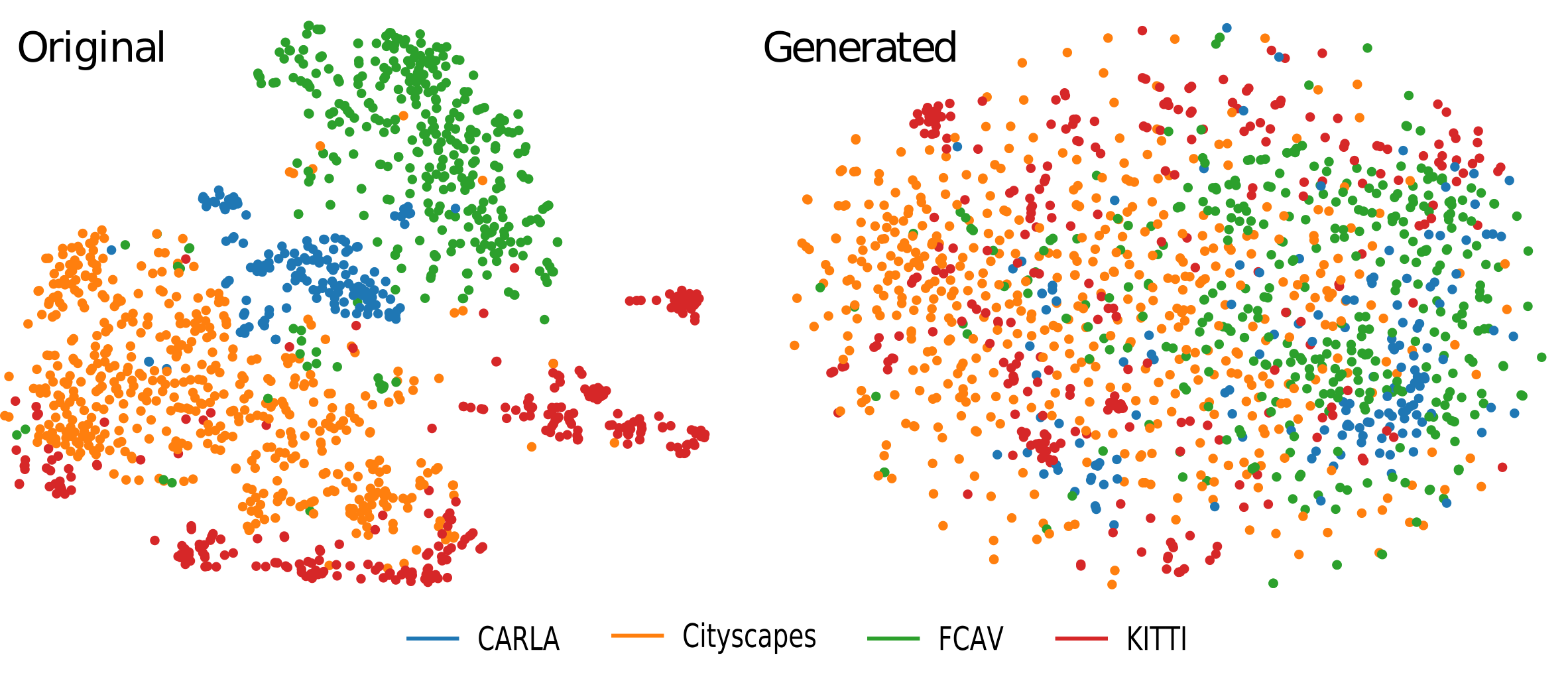}
\vspace{-4mm}
\caption{\label{fig:tsne}t-SNE plots of 1.3k image embeddings of the original images (left) and the generated ones (right).}
\end{center}
\vspace{-14pt}
\end{wrapfigure} 
segmentation degrades faster) and it is unsurprising that our generator tends to prioritize edge information.
In the CARLA example in Figure~\ref{fig:images_examples}, the whole sky is classified as \emph{building} in the EPS map, but the generator learned to ignore such obvious mistakes, as they also occur for the training data.
As another example, the street sign in Figure~\ref{fig:edge} is falsely segmented as \emph{tree}, but correctly recreated as long as the edge information is intact. However, this behaviour makes it harder to guide the image generation process by changing the segmentation map.
In Figure~\ref{fig:segmentation} it can be observed that when the tree line is moved in the segmentation map, trees are still generated according to the edge map (although artifacts appear in the sky).
We expect our method to profit most from improved semantic segmentation networks in the future. 

To validate how well our method is able to overcome the domain disparity present across different data sets we visualize image embeddings generated with a convolutional autoencoder~\cite{10.1145/1390156.1390294} (Figure~\ref{fig:tsne}). We first train the encoder so as to reconstruct the original input images of each data set (left). 
Second, we use the same images, convert them to EPS maps, to then generate photo-realistic images with our pipeline (right). Encoding the original data generates obvious clusters, which indicates that the autoencoder latches on to the unique visual features of each data set (e.g.~desert next to the road for FCAV data, low frequency textures in CARLA, etc.). Images generated with our pipeline are devoid of these artifacts and only show similar visual traits, which results in a more uniform distribution of embeddings. This shows that our method allows us to remove domain shifts that are present in different data sets and for different data modalities.


%% file: 06-conclusion.tex
\smallsection{Conclusion}

We have presented a novel framework that allows us to translate images of various input sources into a unified target domain. Our approach uses EPS maps as an intermediate representation to generate images with photo-realistic, but simplified visual traits. Our goal is to remove common image artifacts, while we maintain enough plausible details to faithfully represent an image. This way, our pipeline can be used to generate training data that facilitates training downstream computer vision tasks, such as classification, object detection, or semantic segmentation. By generating images from EPS maps our method is able to remove a wide range of image artifacts of real data, including seasonal and daytime shifts, camera artifacts such as lens flares, as well as to compensate for common artifacts of synthetic data, such as simplified geometry or low frequency textures.
We have shown that our pipeline is able to generate images with similar visual properties from four different data sources. Furthermore, we have evaluated our approach through a number of ablation studies that show its robustness against common artifacts present across the source domains. However, since the quality of edges is of superior importance compared to segmentations, it is hard to ignore edges impeding the generation of more abstract recreations of the input images. As avenues for future work, it would be interesting to explore the usefulness of our method for an even large number of data sources and for the training of computer vision downstream tasks.